\documentclass[letterpaper]{article} 
\usepackage{aaai2027}  
\usepackage{enumitem}
\usepackage[hyphens]{url}  
\usepackage{graphicx} 

\usepackage{natbib}  
\usepackage{caption} 
\usepackage{subcaption} 
\usepackage{amssymb,amsmath}
\usepackage{mathtools}
\usepackage{pifont} 
\usepackage[export]{adjustbox} 

\usepackage{listings} 
\usepackage{algorithm}
\usepackage{algorithmic}
\usepackage{multirow}
\usepackage{array}
\usepackage{makecell}
\usepackage{booktabs}
\usepackage[table]{xcolor}
\usepackage{arydshln}
\usepackage{xcolor}
\usepackage{amsthm}
\usepackage{tikz}
\usepackage{fvextra}
\usetikzlibrary{positioning,arrows.meta}

\definecolor{delogblue}{HTML}{EAF3FF}
\definecolor{proprietarygray}{gray}{0.93}

\usepackage{newfloat}
\usepackage{listings}
\DeclareCaptionStyle{ruled}{labelfont=normalfont,labelsep=colon,strut=off} 
\floatstyle{ruled}
\newfloat{listing}{tb}{lst}{}
\floatname{listing}{Listing}

\usepackage{xcolor}
\usepackage[most]{tcolorbox}
\tcbuselibrary{listings,breakable}

\newtcblisting{GraphPromptBox}{
    enhanced,
    breakable,
    listing only,
    colback=gray!3,
    colframe=black!65,
    boxrule=0.5pt,
    arc=1mm,
    left=2mm,
    right=2mm,
    top=1mm,
    bottom=1mm,
    title={Prompt for Cognitive Attribution Map Construction},
    colbacktitle=black!65,
    coltitle=white,
    fonttitle=\bfseries\small,
    listing options={
      basicstyle=\ttfamily\scriptsize,
      breaklines=true,
      numbers=none,
      breakatwhitespace=false,
      columns=fullflexible,
      keepspaces=true,
      showstringspaces=false,
      tabsize=2,
      literate=
        {é}{{\'e}}1
        {–}{{--}}1
        {（}{{(}}1
        {）}{{)}}1
    }
  }

  \newtcblisting{AnswerPromptBox}{
    enhanced,
    breakable,
    listing only,
    colback=gray!3,
    colframe=black!65,
    boxrule=0.5pt,
    arc=1mm,
    left=2mm,
    right=2mm,
    top=1mm,
    bottom=1mm,
    title={Prompt for Long-form Answer Generation on the Map},
    colbacktitle=black!65,
    coltitle=white,
    fonttitle=\bfseries\small,
    listing options={
      basicstyle=\ttfamily\scriptsize,
      breaklines=true,
      numbers=none,
      breakatwhitespace=false,
      columns=fullflexible,
      keepspaces=true,
      showstringspaces=false,
      tabsize=2,
      literate=
        {é}{{\'e}}1
        {–}{{--}}1
        {（}{{(}}1
        {）}{{)}}1
    }
  }

\lstdefinelanguage{Datalog}{
  alsoletter={_},
  morekeywords={
    id,name, content, Nodes, Edges,subgraph0, nodes, inputs, outputs, subgraph_description
  },
  morestring=[b]",
  sensitive=true
}
\definecolor{logicblue}{RGB}{54, 112, 178}
\definecolor{logicback}{RGB}{247, 250, 253}
\definecolor{logicframe}{RGB}{166, 190, 216}
\definecolor{logictitle}{RGB}{232, 241, 250}

\usepackage[most]{tcolorbox}
\tcbuselibrary{listings,breakable,skins}

\definecolor{logicback}{RGB}{248,249,250}
\definecolor{logicframe}{RGB}{180,180,180}
\definecolor{logictitle}{RGB}{238,241,245}
\definecolor{logicblue}{RGB}{35,85,140}

\newtcblisting[auto counter]{logicprogram}[2]{
  enhanced,
  breakable,
  listing only,
  colback=logicback,
  colframe=logicframe,
  colbacktitle=logictitle,
  coltitle=black,
  boxrule=0.3pt,
  arc=1pt,
  left=3pt,
  right=3pt,
  top=2pt,
  bottom=2pt,
  boxsep=1pt,
  title={Example graph~\thetcbcounter: #2},
  fonttitle=\bfseries\scriptsize,
  titlerule=0.25pt,
  toptitle=1pt,
  bottomtitle=1pt,
  lefttitle=3pt,
  righttitle=3pt,
  label={#1},
  listing options={
    language=Datalog,
    basicstyle=\fontfamily{pcr}\fontsize{6.8pt}{7.2pt}\selectfont,
    keywordstyle=\color{logicblue!75!black}\bfseries,
    stringstyle=\color{black!75},
    commentstyle=\color{black!55},
    columns=flexible,
    keepspaces=false,
    breaklines=true,
    breakatwhitespace=false,
    breakautoindent=false,
    breakindent=0pt,
    postbreak={},
    showstringspaces=false,
    numbers=none,
    frame=none,
    xleftmargin=0pt,
    aboveskip=0pt,
    belowskip=0pt,
    lineskip=-1pt,
    tabsize=2
  },
  before skip=4pt,
  after skip=4pt
}

\title{Beyond Factual Accuracy: Evaluating Global Reasoning Integrity \\ in RAG Systems with \textsc{LogicScore}}
\author{
Zhichao Yan\textsuperscript{\rm 1},
yunxiao Zhao\textsuperscript{\rm 1},
Jiapu Wang\textsuperscript{\rm 2},
Jiaoyan Chen\textsuperscript{\rm 3},
Xiaoli Li\textsuperscript{\rm 4},
Ru Li\textsuperscript{\rm 1},
Jeff Z. Pan\textsuperscript{\rm 5}
}

\affiliations{
\textsuperscript{\rm 1}Shanxi University, Taiyuan, China\\
\textsuperscript{\rm 2}Nanjing University of Science and Technology, Nanjing, China\\
\textsuperscript{\rm 3}The University of Manchester, Manchester, UK\\
\textsuperscript{\rm 4}Singapore University of Technology and Design, Singapore\\
\textsuperscript{\rm 5}The University of Edinburgh, Edinburgh, UK\\
\texttt{\{yanzhichao,zhaoyunxiao,liru\}@sxu.edu.cn}\\
\texttt{jiapu.wang@njust.edu.cn},
\texttt{jiaoyan.chen@manchester.ac.uk}\\
\texttt{xiaoli\_li@sutd.edu.sg}, \texttt{j.z.pan@ed.ac.uk}
  \vspace{0.3cm} \\ 
  \textbf{Code:} \url{https://github.com/zhichaoyan11/LogicScore}
}

\begin{document}

\maketitle

\begin{abstract}

Current evaluation methods for Retrieval-Augmented Generation (RAG) primarily emphasize factual accuracy, while providing limited assessment of the global reasoning structure of long-form answers. Consequently, RAG systems may produce responses in which individual statements are factually grounded, yet the overall reasoning chain contains missing links, ambiguous connections, or propositions outside the recovered question-to-answer path. To address this limitation, we present \textsc{LogicScore}, a structured diagnostic framework that complements local, fact-by-fact assessment with global reasoning analysis. Using an LLM-assisted, Horn-inspired representation, our approach applies a backward path-recovery mechanism to evaluate three dimensions: \textit{Completeness} (entity-linked path recovery), \textit{Essentiality} (joint path recovery and reasoning density), and \textit{Determinateness} (answer-recovery consistency). Experiments across three multi-hop QA datasets (HotpotQA, MusiQue, and 2WikiMultiHopQA) and more than 20 LLMs (including GPT-5, Gemini-3-Pro, LLaMA3, and task-specific fine-tuned models) show that factual-attribution scores and the proposed structural diagnostics capture complementary properties. Our work provides an inspectable diagnostic framework for reasoning structure, highlighting the importance of considering structural coherence alongside factual grounding in LLM evaluation.

\end{abstract}

\section{Introduction}
Large Language Models (LLMs), pre-trained on massive corpora, have demonstrated remarkable reasoning and generalization capabilities across a broad spectrum of natural language processing tasks \citep{pan2023large,10387715,wang2024large}. Yet, their inherent susceptibility to hallucinations severely undermines reliability, posing a critical barrier to real-world deployment. 
A promising mitigation strategy lies in grounding model outputs in reliable evidence. Attributed Question Answering (AQA) \cite{bohnet2022attributed} enhances trustworthiness and factual accuracy by providing explicit attributions to underlying real-world knowledge sources, enabling users to verify the provenance of generated content.

\begin{figure}[t]
    \centering
    \includegraphics[width=\linewidth]{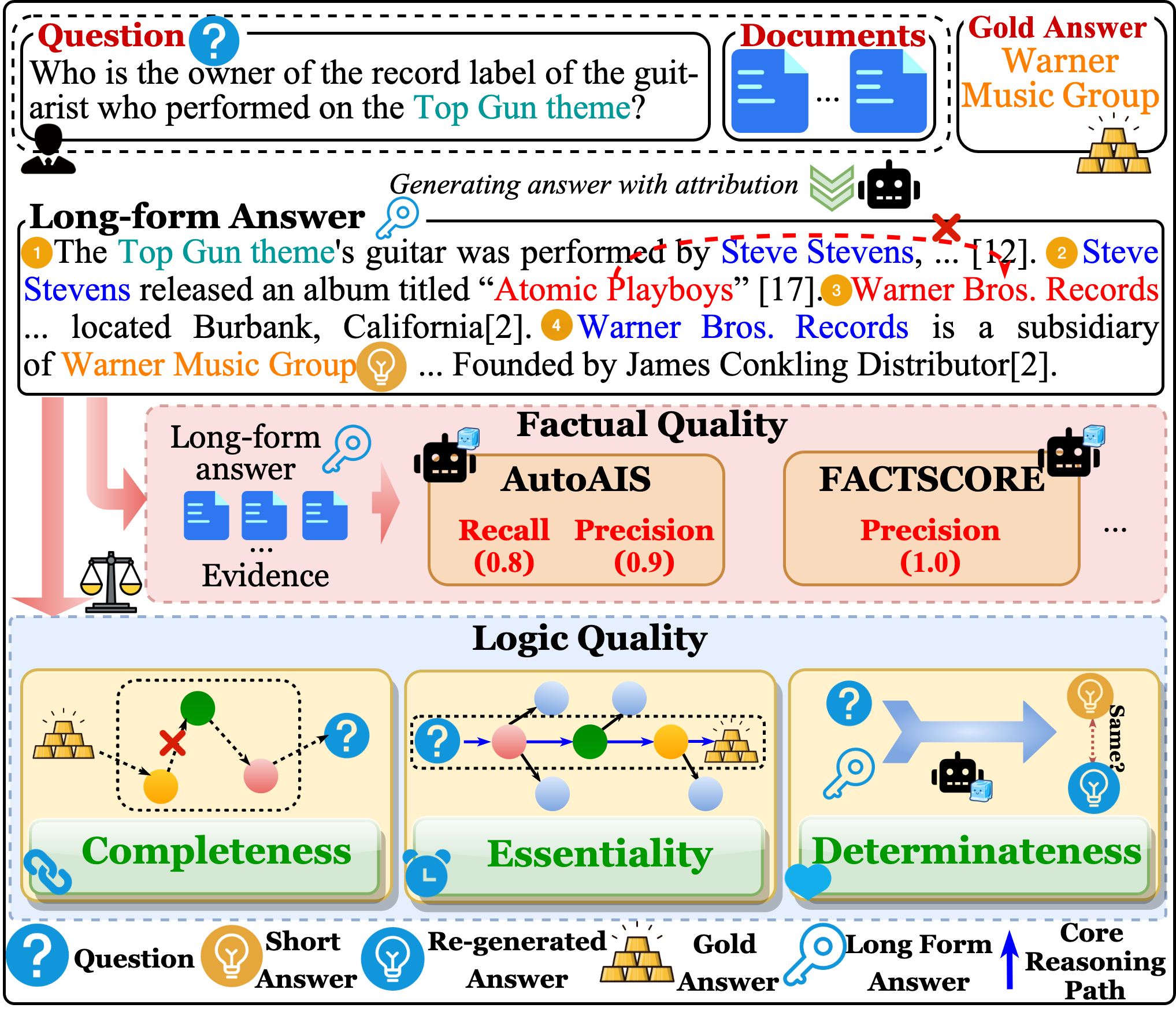}
    \caption{Motivation of \textsc{LogicScore}. Factual evaluation examines whether individual statements are supported by evidence, whereas \textsc{LogicScore} provides complementary diagnostics of the entity-linked reasoning structure through Completeness, Essentiality, and Determinateness.}
    \label{fig:intro}
\end{figure}

\begin{table*}[t]
\centering
\small
\resizebox{\linewidth}{!}{
\begin{tabular}{lccc}
\toprule
\textbf{Method} & \textbf{Verification Unit} & \textbf{Logical Dependency} & \textbf{Reasoning Pattern Identification} \\
\midrule
FACTSCORE \citep{min2023factscore} & Atomic Fact & \ding{55} (Flat) & \ding{55} (Hallucination Only) \\
AutoAIS \citep{gao2023rarr} & Sentence/Attribution & \ding{55} (Flat) & \ding{55} (Hallucination Only) \\
CoT Evaluation & Unstructured Text & $\approx$ (Implicit) & Weak (Often misses circularity) \\
\midrule
\rowcolor{gray!15} \textbf{\textsc{LogicScore} (Ours)} & \textbf{Reasoning Chain} & \textbf{\ding{51} (Horn Clauses)} & \textbf{\ding{51} (Circular, Broken, Deviated)} \\
\bottomrule
\end{tabular}}
\caption{\textsc{LogicScore} differs from existing paradigms by evaluating logical construction beyond factual verification. \textit{Verification Unit} indicates the evaluation granularity, \textit{Logical Dependency} captures modeling of logical dependencies, and \textit{Reasoning Pattern Identification} measures the ability to identify invalid reasoning patterns.}
\label{tab:comparison}

\end{table*}

Previous RAG research generally focuses on evaluating contextual faithfulness, primarily quantifying how accurately the generated responses reflect the retrieved facts \citep{rashkin2023measuring,gao2023rarr,yan2025atomic,hirsch2025laquerlocalizedattributionqueries,zhang2025longcite,yan2025decomposing}. However, this factuality-first evaluation paradigm (e.g., AutoAIS \citep{gao2023rarr} and FactScore \citep{min2023factscore}) alone may provide an incomplete account of long-form answer quality. Optimizing factual support at the level of individual statements does not necessarily ensure that these statements form a coherent question-to-answer reasoning structure. Many widely used factuality metrics (factual quality metrics in \textcolor[HTML]{e49b97}{pink} Figure \ref{fig:intro}) primarily evaluate whether individual statements are supported by evidence, while providing limited information about how these statements collectively connect the question to the answer. We refer to this limitation as \textit{factual myopia}. Consequently, a long-form answer may contain individually supported statements while still omitting an intermediate link, introducing an ambiguous connection, or including propositions that do not participate in the recovered question-to-answer path. Figure \ref{fig:intro} illustrates this distinction. Although most statements in the example are individually supported, the generated explanation does not establish an explicit connection between \texttt{Steve Stevens} and \texttt{Warner Bros. Records}. This issue is further amplified by the complex interplay of cross-document facts, which obscures intrinsic logical relationships. This motivates complementary evaluation methods that examine the organization of reasoning steps alongside factual support.

Inspired by Hegelian Logic \citep{hegel1975hegel}, we formalize global evaluation beyond ad-hoc heuristics using three principles of ideal reasoning: Essence (the core of logical structure), Totality (gap-free deductive coverage), and Determinateness (answer-recovery consistency of conclusions). To operationalize these dimensions, we organize the extracted propositions using a Horn-inspired representation \citep{levy1998combining}. This representation makes the candidate question-to-answer path explicit and exposes intermediate evaluation decisions for inspection \citep{ferrand2005explanations}. It should be understood as an operational abstraction of the generated reasoning structure rather than a complete formalization of natural-language semantics.

Building on this motivation, we propose \textsc{LogicScore}, an LLM-assisted structured diagnostic framework for entity-centric multi-hop answers, which 
systematically assesses three complementary dimensions of global reasoning quality (Figure \ref{fig:intro}, \textcolor[HTML]{86a5db}{blue} area), involving: 
(1) \textbf{Completeness} indicates whether the extracted propositions contain an entity-linked path connecting the question to the gold answer, aligning with Hegel’s ``Totality''; (2) \textbf{Essentiality} measures the proportion of propositions participating in the recovered path and is set to zero when no complete path is found, thereby jointly reflecting path recovery and reasoning density, reflecting ``Essence'';  (3) \textbf{Determinateness} measures whether an evaluator can recover the original short answer from the question and long-form answer alone, embodying ``Determinateness''. 
Specifically, \textit{Completeness} is calculated by checking if a valid reasoning path can be constructed from the gold answer to the question (scored as 1 for success, 0 for failure). Built on \textit{Completeness}, we further calculate the \textit{Essentiality} as the ratio of the number of steps in the valid path to the total number of steps in the long-form answer. \textit{Determinateness} is computed by testing whether the long-form answer successfully entails the short answer.

Equipped with these quantitative measures, we examine \textsc{LogicScore} across three knowledge-intensive multi-hop QA benchmarks (HotpotQA, MusiQue, and 2WikiMultiHopQA) and more than 20 LLMs, including proprietary models (e.g., GPT-5.1 and Gemini-3-Pro), open-source models (e.g., LLaMA-3 and Qwen3), and task-specific fine-tuned models. The results show that factual-attribution scores and the proposed structural diagnostics can yield substantially different evaluation profiles. For example, Gemini-3-Pro obtains an attribution precision of 92.85\% while receiving an Essentiality score of 35.11\% under the path-based definition used in this work. We further observe that the structural diagnostics do not increase monotonically with model size, indicating that answer-recovery consistency, path recovery, and path density capture distinct properties of generated responses.

Taken together, this work provides a complementary diagnostic perspective on the reasoning structure of evidence-grounded, entity-centric multi-hop answers alongside factual grounding. By exposing propositions, extracted triples, recovered paths, and re-inferred answers as intermediate artifacts, \textsc{LogicScore} makes its structural assessments more inspectable within this task setting.

\section{Related Work}

\textbf{Attributed Question Answering.}
Recent AQA methods broadly follow Retrieval-then-Read or Post-hoc Retrieval paradigms. Retrieval-then-Read retrieves top-$k$ documents and generates attributed answers through CoT prompting, supervised fine-tuning, or preference optimization \citep{gao-etal-2023-enabling,aly2024learning,huang2024learning,li2024improving}; Post-hoc Retrieval generates answers first and subsequently retrieves evidence for attribution or verification \citep{gao2023rarr,chen2024complex,slobodkin2024attribute,yan2025atomic}. Knowledge graphs have also supported AQA evaluation-resource construction \citep{hu-etal-2025-llms}. \textsc{LogicScore} diagnoses how propositions in attributed long-form answers form an entity-linked question-to-answer path.

\textbf{Factual Evaluation.}
AIS, AutoAIS, ALCE, and subsequent LLM-based evaluators assess whether generated content is supported by identifiable evidence or citations \citep{rashkin2023measuring,gao2023rarr,gao-etal-2023-enabling,zhang2025longcite,xu2025citeeval,rawte2025document}. These metrics measure evidence support and faithfulness; \textsc{LogicScore} is complementary in examining the recovered structure of a reasoning path, rather than verifying factual support itself.

\textbf{Chain-of-Thought Evaluation.}
Automatic CoT evaluation has moved beyond final-answer accuracy. ROSCOE and ReCEval assess properties including step correctness, informativeness, consistency, and coherence \citep{golovneva2023roscoe,prasad2023receval}. Direct Evaluation evaluates multi-hop CoT against ground-truth knowledge graphs, SocREval performs reference-free LLM evaluation, and REVEAL provides step-level annotations for reasoning verifiers \citep{nguyen2024direct,he2024socreval,jacovi2024chain}. In contrast, \textsc{LogicScore} constructs an answer-specific entity-linked representation for attributed long-form answers and reports path-level structural diagnostics without requiring a reference reasoning chain or ground-truth reasoning graph \ref{tab:comparison}; its representation is, however, LLM-generated and therefore subject to transformation uncertainty.

\begin{figure*}[t]
    \centering
    \includegraphics[width=1\linewidth]{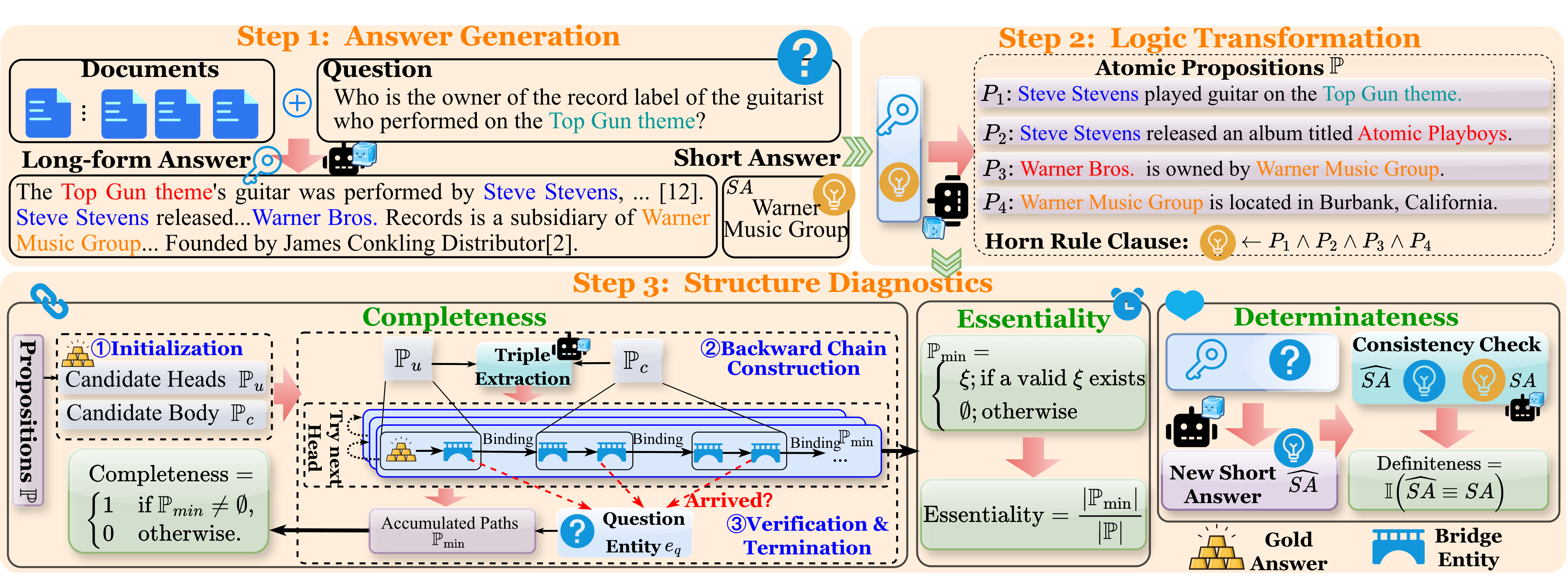}
    \caption{Overview of the \textsc{LogicScore} evaluation framework, consisting of three phases: (1) Answer Generation, where the model produces a Long-form Answer (\(\mathcal{LA}\)) and a Short Answer (\(\mathcal{SA}\)) based on the question and top-$k$ documents; (2) Logic Transformation, which decomposes the $\mathcal{LA}$ into a set of atomic propositions ($\mathbb{P}$) structured as Horn clauses; and (3) Structure Diagnostics, which assesses the reasoning quality across three dimensions: Completeness, Essentiality and Determinateness.}
    \label{fig:framework}
\end{figure*}

\section{Preliminaries}

In this section, we introduce the AQA setting, notation, and the Horn-inspired representation used to organize entity-linked reasoning structures \citep{bohnet2022attributed,gao-etal-2023-enabling,gao2023rarr}.

\noindent\textbf{Attributed Question Answering.}
In the setting considered in this work, an AQA model generates a long-form answer with attributions and a corresponding short answer. Given a question $\mathcal{Q}$ and a set of retrieved passages $\mathcal{D}$, we write:
\begin{equation}
    (\mathcal{LA},\mathcal{SA})
    \longleftarrow
    \mathcal{M}_{\textsc{AQA}}(\mathcal{Q},\mathcal{D}),
    \nonumber
\end{equation}
where $\mathcal{LA}$ denotes the attributed long-form answer, $\mathcal{SA}$ denotes its short answer, and $\mathcal{M}_{\textsc{AQA}}$ is the evaluated AQA model. We use \textit{attribution} as a general term covering both citations generated in Retrieval-then-Read settings \citep{gao-etal-2023-enabling} and evidence retrieved for post-hoc verification \citep{yan2025atomic}.

\noindent\textbf{Structural Diagnostics.}
We consider three operational dimensions: \textit{Completeness} tests whether the extracted propositions contain an entity-linked path between the question and the gold answer; \textit{Essentiality} measures the proportion of propositions participating in the recovered path and is set to zero when no complete path is found; and \textit{Determinateness} measures whether the original short answer can be recovered from the question and long-form answer through LLM-based re-inference.

\noindent\textbf{Definite Horn Rules and Operational Path Representation.}
A definite Horn rule is a restricted logical expression consisting of a conjunction of body atoms and a single positive head atom \citep{cheng2021uniker}. It can be written as:
\begin{equation}
    r_0(x,y)
    \leftarrow
    r_1(x,z_1)\wedge\cdots\wedge r_n(z_{n-1},y),
\end{equation}
where $r_0(x,y)$ is the rule head and the sequence of atoms $r_i(\cdot)$ forms the rule body.

Inspired by this body--head organization, \textsc{LogicScore} uses extracted natural-language propositions as candidate support for an answer and represents their structural connection as an entity--relation path:
\begin{equation}
    \pi:
    e_{\mathcal{Q}}
    \xrightarrow{r_1}
    e_1
    \xrightarrow{r_2}
    \cdots
    \xrightarrow{r_k}
    e_{\mathrm{gold}},
    \label{eqh}
\end{equation}
where $e_{\mathcal{Q}}$ is an entity identified in the question, $e_{\mathrm{gold}}$ is the gold-answer entity, and each relation edge is induced by an extracted proposition \(P\).

For example, the propositions ``Steve Stevens played guitar on the Top Gun theme'' and ``Steve Stevens released an album titled Atomic Playboys'' induce the path
\texttt{Top Gun theme $\xrightarrow{\text{performed\_by}}$ Steve Stevens $\xrightarrow{\text{released}}$ Atomic Playboys}.
This path makes shared-entity connectivity explicit and inspectable; it does not by itself establish that one proposition formally entails the next. The path traversal is deterministic once the propositions and triples are fixed, whereas construction of those intermediate representations remains LLM-dependent.

\section{Methodology}

This section presents the three stages of \textsc{LogicScore}: \textit{Stage 1: Answer Generation} prompts an LLM to produce an attributed long-form answer and a short answer; \textit{Stage 2: Logic Transformation} decomposes the long-form answer into atomic propositions and construct an answer-specific structured representation; and \textit{Stage 3: Structural Diagnostics} recovers an entity-linked path from this representation and computes Completeness, Essentiality, and Determinateness.

\textbf{Answer Generation.} Given a query $\mathcal{Q}$ and retrieved documents $\mathcal{D}$, our objective is to produce a detailed long-form response ($\mathcal{LA}$) with citations, followed by a concise short answer ($\mathcal{SA}$). To elicit explicit intermediate steps, we use a CoT prompting strategy \citep{Ji_Liu_Du_Ng_2024}. The model is prompted to produce a step-by-step explanation, used as $\mathcal{LA}$, with attribution symbols attached to the corresponding statements, followed by the final $\mathcal{SA}$. Formally, the whole process can be represented as:
\begin{tcolorbox}[colback=blue!1!white,colframe=blue!65!green, title=\textit{Who is the owner of the record label of the guitarist who performed on the Top Gun theme?}]
\textbf{Long-form Answer $\mathcal{LA}$:} \textit{The Top Gun theme's guitar was performed by Steve Stevens, ... [12]. Steve Stevens released an album titled “Atomic Playboys” [17]. Warner Bros. Records ... located Burbank, California [2]. Warner Bros. Records is a subsidiary of Warner Music Group ... Founded by James Conkling Distributor  [2].}

\textbf{Short Answer $\mathcal{SA}$:} \textit{Warner Music Group}
\end{tcolorbox}
\noindent Note we use some ellipses to indicate that words have been omitted due to space limitations, and \([\text{num}]\) denotes a citation where num represents the document number in \(\mathcal{D}\).  

\textbf{Logic Transformation.}
Long-form answers may contain multiple intertwined statements, making their question-to-answer structure difficult to inspect directly. We therefore use an LLM to decompose $\mathcal{LA}$ into a set of atomic natural-language propositions $\mathbb{P}=\{P_1,\ldots,P_n\}$. The propositions are organized as an answer-specific, Horn-inspired support representation associated with $\mathcal{SA}$. The transformation is written as:
\begin{equation}
    \mathcal{T}(\mathcal{LA},\mathcal{SA})
    \longrightarrow
    (\mathbb{P},\mathcal{SA}).
\end{equation}
Here, the propositions are treated as candidate support units rather than formal logical atoms that prove the short answer. Because $\mathcal{T}$ is implemented using an LLM, the resulting representation may be affected by proposition segmentation, entity identification, and relation-extraction errors.

\textbf{Structural Diagnostics.}
Given the transformed proposition set, \textsc{LogicScore} computes three operational diagnostics of the recovered reasoning structure. These diagnostics characterize the representation constructed by the pipeline rather than establishing formal logical validity.

(1) \textit{Completeness.}
Completeness is operationalized as a gold-answer-anchored path-recovery test over the extracted representation. Starting from the gold-answer entity $e_\text{gold}$, the procedure searches backward for a sequence of propositions connected to an entity $e_{\mathcal{Q}}$ identified in the question. Here, $e_\text{gold}$ is obtained from the benchmark reference answer and is distinct from the model-generated short answer $\mathcal{SA}$. The procedure consists of three stages:

\textbf{Candidate Initialization.}
Given $\mathbb{P}=\{P_k\}_{k=1}^{n}$, we identify the subset
$\mathbb{P}_u=\{P_i\}_{i=1}^{m}$ containing $e_\text{gold}$. These propositions serve as starting candidates for backward search. The remaining propositions form the context set $\mathbb{P}_c$.

\textbf{Backward Path Construction.}
Starting from each candidate in $\mathbb{P}_u$, the procedure recursively searches for connected propositions. An evaluator LLM parses each visited proposition into an entity--relation triple $(e_{sub},rel,e_{obj})$. Given a current entity $e_{bridge}$, the search scans $\mathbb{P}_c$ for a proposition in which the same entity occurs as an argument and then updates $e_{bridge}$ using the other entity in the matched triple. The primary matching criterion is therefore shared-entity connectivity. Relation labels are retained in the recovered representation for inspection, but this matching operation does not by itself establish the formal compositional validity of adjacent predicates.

\textbf{Path Termination.}
The search succeeds when it reaches a proposition containing $e_{\mathcal{Q}}$. The resulting sequence
\[
    \xi=\{P_{\xi_1},\ldots,P_{\xi_k}\}
\]
is denoted as the recovered path set $\mathbb{P}_{path}$. If the search space is exhausted without reaching $e_{\mathcal{Q}}$, we set $\mathbb{P}_{path}=\emptyset$.

Completeness is then defined as:
\begin{equation}
    \mathrm{Completeness} =
    \begin{cases}
    1 & \text{if } \mathbb{P}_{path}\neq\emptyset,\\
    0 & \text{otherwise}.
    \end{cases}
\end{equation}
In Figure \ref{fig:framework}, $P_1$ connects to $P_2$ through ``Steve Stevens'' but the search does not reach the question entity. Thus, $\mathbb{P}_{path}=\emptyset$ and Completeness is 0. Because the framework does not assume a unique gold reasoning path, Completeness is reported as a binary path-recovery indicator. It should not be interpreted as a formal proof that the recovered path is semantically valid.

\begin{table*}[t]
\centering
\small
\setlength{\tabcolsep}{0.15cm}{%
\begin{tabular}{lcccccccccc}
\toprule
\multirow{2}{*}{Models} & \multicolumn{1}{c|}{\multirow{2}{*}{Size}} & \multicolumn{3}{c|}{\textbf{HotpotQA}} & \multicolumn{3}{c|}{\textbf{MusiQue}} & \multicolumn{3}{c}{\textbf{2WikiMultiHopQA}} \\ \cmidrule(r{6pt}l{6pt}){3-5} \cmidrule(r{6pt}l{6pt}){6-8} \cmidrule(r{6pt}l{6pt}){9-11}
 & \multicolumn{1}{c|}{} & Esse. & Comp. & \multicolumn{1}{c|}{Dete.} & Esse. & Comp. & \multicolumn{1}{c|}{Dete.} & Esse. & Comp. & Dete. \\
\multicolumn{11}{c}{\cellcolor[HTML]{fbe7cf}\textbf{Proprietary LLMs}} \\
GPT-4o & \multicolumn{1}{c|}{UNK} & 44.90 & 75.57 & \multicolumn{1}{c|}{97.36} & 32.54 & 57.10 & \multicolumn{1}{c|}{93.66} & 46.55 & 80.19 & 93.13 \\
GPT-o3 &  \multicolumn{1}{c|}{UNK} & 47.44 & 73.24 &  \multicolumn{1}{c|}{\underline{98.92}} & \underline{37.30} & \underline{60.83} &  \multicolumn{1}{c|}{\textbf{96.48}} & 50.00 & 82.37 & \textbf{95.80} \\
GPT-5.1 & \multicolumn{1}{c|}{UNK} & \underline{48.37} & 72.61 & \multicolumn{1}{c|}{\textbf{99.15}} & 27.27 &  60.11 & \multicolumn{1}{c|}{87.79} & 37.96 & 81.00 & 92.45 \\
Gemini-3-pro & \multicolumn{1}{c|}{UNK} & 39.69 & \textbf{78.65} & \multicolumn{1}{c|}{98.56} &  \textbf{38.98} & 59.21 & \multicolumn{1}{c|}{\underline{94.10}} & \underline{53.93} & \underline{85.78} & \underline{94.69} \\
Claude-4.5 & \multicolumn{1}{c|}{UNK} & 36.40 & \underline{78.33} & \multicolumn{1}{c|}{98.07} & 27.13 & \textbf{63.88} & \multicolumn{1}{c|}{89.66} & 40.13 &  \textbf{85.86} & 92.03 \\
DeepSeek-V3.2 & \multicolumn{1}{c|}{671B} & 44.81 & 76.34 & \multicolumn{1}{c|}{97.96} & 30.34 & 59.50 & \multicolumn{1}{c|}{86.33} & 44.32 & 79.49 & 90.89 \\
DeepSeek-R1 &  \multicolumn{1}{c|}{671B} & \textbf{49.60} & 73.91 &  \multicolumn{1}{c|}{98.56} & 34.33 & 55.82 &  \multicolumn{1}{c|}{91.12} & \textbf{56.41} & 84.16 & 93.13 \\
\multicolumn{11}{c}{\cellcolor[HTML]{ccccfb}\textbf{Open-source LLMs}} \\
 & \multicolumn{1}{c|}{1B} & 10.39 & 17.45 & \multicolumn{1}{c|}{46.45} & 2.37 & 3.82 & \multicolumn{1}{c|}{26.03} & 12.73 & 18.55 & 44.85 \\
\multirow{-2}{*}{LLaMA-3.2} & \multicolumn{1}{c|}{3B} & 10.39 & 19.02 & \multicolumn{1}{c|}{70.41} & 3.46 & 7.43 & \multicolumn{1}{c|}{55.42} & 28.37 & 42.00 & 58.45 \\ \cdashline{2-11}
\rule{0pt}{9pt} & \multicolumn{1}{c|}{8B} & 26.46 & 54.94 & \multicolumn{1}{c|}{84.28} & 12.35 & 28.55 & \multicolumn{1}{c|}{65.96} & 28.65 & 60.22 & 69.70 \\
\multirow{-2}{*}{LLaMA-3.1} & \multicolumn{1}{c|}{70B} & 39.40 & 71.40 & \multicolumn{1}{c|}{96.04} & 27.72 & 51.92 & \multicolumn{1}{c|}{86.06} & 43.79 & 75.62 & 90.48 \\ \cdashline{2-11}
 \rule{0pt}{9pt} & \multicolumn{1}{c|}{0.6B} & 33.11 & 52.16 & \multicolumn{1}{c|}{81.52} & 6.92 & 11.32 & \multicolumn{1}{c|}{58.12} & 31.53 & 49.78 & 64.03 \\
 & \multicolumn{1}{c|}{1.7B} & 50.07 & 66.17 & \multicolumn{1}{c|}{92.90} & 21.56 & 30.39 & \multicolumn{1}{c|}{73.05} & 52.06 & 71.64 & 82.53 \\
 & \multicolumn{1}{c|}{4B} & 50.39 & \underline{75.58} & \multicolumn{1}{c|}{96.87} & 29.83 & 46.88 & \multicolumn{1}{c|}{84.40} & 53.85 & 81.53 & 91.23 \\
 & \multicolumn{1}{c|}{8B} & 47.47 & 75.36 & \multicolumn{1}{c|}{97.68} & 31.33 & 52.06 & \multicolumn{1}{c|}{\underline{86.71}} & 52.88 & 81.33 & 90.95 \\
 & \multicolumn{1}{c|}{14B} & \textbf{52.01} & 73.18 & \multicolumn{1}{c|}{97.31} & 31.78 & 48.81 & \multicolumn{1}{c|}{85.60} & \underline{54.56} & 81.20 & 90.55 \\
 & \multicolumn{1}{c|}{30B-A3B} & \underline{50.97} & \textbf{77.09} & \multicolumn{1}{c|}{\underline{97.77}} & \underline{32.41} & \underline{53.69} & \multicolumn{1}{c|}{85.06} & \textbf{55.99} & \underline{82.73} & \underline{91.86} \\
\multirow{-7}{*}{Qwen3} & \multicolumn{1}{c|}{235B-A22B} & 45.62 & 73.80 & \multicolumn{1}{c|}{\textbf{98.87}} & \textbf{35.25} & \textbf{60.94} & \multicolumn{1}{c|}{\textbf{95.62}} & 53.25 & \textbf{86.00} & \textbf{96.74} \\
\multicolumn{11}{c}{\cellcolor[HTML]{ecced9}\textbf{Supervised Fine-tuning LLMs}} \\
FRONT & \multicolumn{1}{c|}{13B} & 27.76 & 57.96 & \multicolumn{1}{c|}{82.86} & 6.34 & 13.70 & \multicolumn{1}{c|}{61.19} & 26.08 & 50.23 & 69.94 \\
LongCite & \multicolumn{1}{c|}{8B} & \underline{38.85} & \underline{64.56} & \multicolumn{1}{c|}{\textbf{91.75}} & \underline{17.55} & \underline{32.73} & \multicolumn{1}{c|}{\textbf{76.04}} & \textbf{40.13} & \textbf{67.88} & \textbf{72.30} \\
SelfCite & \multicolumn{1}{c|}{8B} & \textbf{39.23} & \textbf{66.74} & \multicolumn{1}{c|}{\underline{91.04}} & \textbf{19.02} & \textbf{35.16} & \multicolumn{1}{c|}{\underline{75.59}} & \underline{39.22} & \underline{66.47} & \underline{72.17}  \\ \bottomrule
\end{tabular}%
\caption{Evaluation results on three datasets across three types of LLMs. ``Esse.'', ``Comp.'' and ``Dete.'' represent ``Essentiality'', ``Completeness'' and ``Determinateness'', respectively. The best two results within each LLM type are \textbf{bold} and \underline{underlined}.}
\label{tab:main}
}
\end{table*}

(2) \textit{Essentiality.}
This metric summarizes both successful path recovery and the proportion of extracted propositions participating in the recovered path. Using the path set defined above, we compute:
\begin{equation}
    \mathrm{Essentiality}
    =
    \frac{\lvert\mathbb{P}_{path}\rvert}
         {\lvert\mathbb{P}\rvert},
\end{equation}
where $\mathbb{P}_{path}=\emptyset$ when no complete question-to-answer path is recovered. Consequently, Essentiality is 0 whenever Completeness is 0. When Completeness is 1, the metric measures the proportion of extracted propositions included in the recovered path.

Essentiality is therefore a joint path-recovery and density metric rather than a pure measure of redundancy. A low score may reflect failure to recover a complete path, the presence of propositions outside the selected path, or errors in the automatically constructed representation. For example, the failed chain $\mathbb{P}_{f}$ in Figure \ref{fig:framework} has $\mathbb{P}_{path}=\emptyset$ and therefore receives an Essentiality score of 0.

(3) \textit{Determinateness.}
We operationalize Determinateness through an LLM-based answer re-inference procedure. Given $\mathcal{Q}$ and $\mathcal{LA}$, an evaluator model $\mathcal{M}$ generates a new short answer $\hat{\mathcal{SA}}$:
\begin{equation}
    \hat{\mathcal{SA}}
    \sim
    \mathcal{M}(\cdot\mid\mathcal{Q},\mathcal{LA}).
\end{equation}
The retrieved documents $\mathcal{D}$ are not supplied during this step, making $\mathcal{LA}$ the only explicit evidential context provided to the evaluator. This restriction reduces access to the retrieved evidence but does not prevent the evaluator from drawing on its parametric knowledge.

Determinateness is calculated using the adopted answer-matching criterion:
\begin{equation}
    \mathrm{Determinateness}
    =
    \mathbb{I}
    \bigl(
        \mathrm{Match}(\hat{\mathcal{SA}},\mathcal{SA})=1
    \bigr),
\end{equation}
where $\mathrm{Match}(\cdot,\cdot)$ determines whether the two short answers are equivalent under the answer-normalization procedure, and $\mathbb{I}(\cdot)$ is the indicator function. A score of 1 indicates that the original short answer is recovered under this re-inference procedure, whereas 0 indicates an answer mismatch. The score measures answer-recovery consistency and should not be interpreted as a formal proof that $\mathcal{LA}$ uniquely or strictly entails $\mathcal{SA}$.

The three diagnostics describe complementary but operationally related properties of the generated answer. In particular, Essentiality is definitionally conditioned on path recovery and should therefore be interpreted together with Completeness. Determinateness separately measures whether the original short answer can be recovered from the long-form answer under the evaluator model. Their conceptual relation to Hegelian Logic is discussed in the Appendix; the metric definitions themselves are operational and do not constitute a formal logical system.

\section{Experiments}

\label{main_results}
\subsection{Experiment Setups}

\label{setup}

\noindent \textbf{Datasets.} Motivated by the need to assess complex logical reasoning on knowledge-intensive QA task, we experiment on three challenging multi-hop datasets: \textit{HotpotQA} \citep{yang2018hotpotqa}, \textit{MusiQue} \citep{trivedi2022musique}, and \textit{2WikiMultiHopQA} \citep{ho2020constructing}. To isolate structure diagnostics from retrieval errors, we utilize the ``distractor'' setting in HotpotQA and the ``answerable'' setting in MusiQue. We further evaluate LogicScore on LLM-selected subsets requiring chain-structured multi-hop reasoning, comprising 1,939, 1,605, and 2,177 instances from HotpotQA, MusiQue, and 2WikiMultiHopQA, respectively. The selection prompt is provided in Appendix.

\noindent \textbf{LLMs.} To ensure comprehensive evaluation, we assess proprietary (\texttt{GPT} series, \texttt{Gemini-3-Pro}, \texttt{Claude-4.5}, \texttt{DeepSeek} series), open-source (\texttt{LLaMA3} and \texttt{Qwen3} series), and task-specific SFT models (\texttt{FRONT} (LLaMA2-13B) \citep{huang2024learning}, \texttt{LongCite} (LLaMA3-8B) \citep{zhang2025longcite}, and \texttt{SelfCite} (LLaMA3-8B) \citep{chuangselfcite}). \textsc{LogicScore} utilizes GPT-4o-mini for \textit{Completeness} and \textit{Essentiality}, and GPT-o4-mini for \textit{Determinateness}. Implementation, factual evaluation, and prompt details are in Appendix.

\subsection{Main Results and Analysis}

Table \ref{tab:main} reports the three structural diagnostics across models and datasets. As the main results are based on a single run per configuration, we interpret the following comparisons as descriptive trends.

(1) Proprietary models generally achieve high Determinateness but lower Completeness and Essentiality on MusiQue. For example, Gemini-3-Pro and GPT-o3 obtain Determinateness scores of 94.10\% and 96.48\%, while Gemini-3-Pro achieves 59.21\% Completeness and GPT-o3 achieves 37.30\% Essentiality. This suggests that short answers are often recoverable even when a complete entity-linked path is not consistently recovered. Since Essentiality is conditioned on Completeness, its lower value should not be interpreted solely as redundancy.

(2) Open-source models exhibit non-monotonic patterns across model sizes. On HotpotQA, Qwen3-235B-A22B achieves 98.87\% Determinateness but 45.62\% Essentiality, compared with 52.01\% for Qwen3-14B. A similar difference appears on MusiQue. These results indicate that answer-recovery consistency and path-recovery density capture different structural properties, but do not establish that scaling causes greater verbosity or redundancy.

(3) SFT models show dataset-dependent results. SelfCite (8B) achieves 66.74\% Completeness on HotpotQA, compared with 54.94\% for LLaMA-3.1-8B, but decreases to 35.16\% on MusiQue. This variation may reflect differences in task difficulty, generation format, reasoning structure, or representation quality. Additional analyses across reasoning depths and model sizes are provided in the Appendix.

\begin{table}[t] 
\centering
\small
\setlength{\tabcolsep}{2pt}{
\begin{tabular}{llccc} 
\toprule
Methods & \multicolumn{1}{c}{Models} & Esse. & Comp. & Dete. \\
\midrule
\multirow{3}{*}{Vanilla} & GPT-5.1 & 24.16 & 51.79 & 88.25 \\
 & Gemini-3-Pro & 32.33 & 57.96 & 95.30 \\
 & Qwen3-235B-A22B & 23.24 & 54.21 & 92.71 \\ 
\midrule
\multirow{3}{*}{CoT} & GPT-5.1 & 27.27$_{\textcolor{red}{\uparrow 3.11}}$ & 60.11$_{\textcolor{red}{\uparrow 8.32}}$ & 87.79$_{\textcolor{green}{\downarrow 0.46}}$ \\
 & Gemini-3-Pro & 38.98$_{\textcolor{red}{\uparrow 6.75}}$ & 59.21$_{\textcolor{red}{\uparrow 1.24}}$ & 94.10$_{\textcolor{green}{\downarrow 1.20}}$ \\
 & Qwen3-235B-A22B & 35.25$_{\textcolor{red}{\uparrow 12.01}}$ & 60.94$_{\textcolor{red}{\uparrow 6.73}}$ & 95.62$_{\textcolor{red}{\uparrow 3.91}}$ \\ 
 \bottomrule
\end{tabular}%
}
\caption{Influence of different prompt strategies for long-form answer generation on the MusiQue dataset. We use the prompt template proposed by \citet{gao-etal-2023-enabling} as the Vanilla method and CoT template from \citet{ji2024chain}.} 
\label{tab:prompt}
\end{table}

\subsection{Robustness and Reliability Analysis}

\textbf{Sensitivity to Answer-Generation Prompts.} Table \ref{tab:prompt} shows that CoT prompting improves Completeness and Essentiality over the Vanilla baseline for the evaluated models on MusiQue. For Qwen3-235B-A22B, Essentiality increases from 23.24\% to 35.25\%, accompanied by a 6.73\% gain in Completeness. These results indicate that the proposed diagnostics are responsive to changes in answer-generation prompts, although Determinateness varies across models.

\textbf{Human Evaluation and Transformation Quality.}
We compare \textsc{LogicScore} with direct LLM-as-a-Judge evaluation on 100 randomly sampled instances. As shown in Table \ref{tab:human_logicscore_llm_judge}, \textsc{LogicScore} achieves higher agreement with human judgments in all six dataset--metric comparisons. Averaged across the two datasets, it outperforms LLM-as-a-Judge by 17.57\% for Essentiality, 11.66 for Completeness, and 4.44\% for Determinateness. The larger gains on Essentiality and Completeness suggest that proposition- and path-level evaluation better reflects human assessments of reasoning efficiency and connectivity than direct end-to-end judging.

Separately, the Stage 2 logic transformation achieves human-evaluated accuracies of 94.14\% on MusiQue and 97.89\% on 2WikiMultiHopQA, indicating high semantic fidelity. Overall, \textsc{LogicScore} provides more human-aligned evaluations while maintaining reliable intermediate representations. Annotation criteria, prompts, and metric calculations are provided in the Appendix.

\textbf{Consistency across Evaluator Backbones.}
Table \ref{tab:differe_llm} reports Cohen's Kappa agreement when the evaluator backbone is changed. Most scores exceed 70\%, indicating substantial agreement, while Determinateness exceeds 90\% across the evaluated settings. This suggests that the diagnostics are relatively consistent across the tested evaluator backbones, although the pipeline remains dependent on LLM-generated intermediate representations.

\begin{table}[t]
  \centering
  \small
  \setlength{\tabcolsep}{2pt}
  \begin{tabular}{llcccc}
  \toprule
  \multirow{2}{*}{Datasets} & \multirow{2}{*}{Evaluator} & Pearson-$r\uparrow$ & \multicolumn{2}{c}{Jaccard$\uparrow$} & Stage 2 \\
  \cmidrule(lr){3-3} \cmidrule(lr){4-5} \cmidrule(lr){6-6}
   & & Esse. & Comp. & Dete. & Acc.$\uparrow$ \\
  \midrule
  \multirow{2}{*}{MusiQue}
      & LogicScore       & 86.32 & 91.48 & 90.81 & 94.14 \\
      & LLM-as-a-Judge   & 73.57 & 84.21 & 83.16 & -- \\
  \midrule
  \multirow{2}{*}{2Wiki}
      & LogicScore       & 93.77 & 97.87 & 93.39 & 97.89 \\
      & LLM-as-a-Judge   & 71.39 & 81.82 & 92.16 & -- \\
  \bottomrule
  \end{tabular}
  
  \caption{Human evaluation of metric correlations and transformation accuracy for LogicScore and pure LLM-as-a-Judge. Pearson-$r$ measures agreement on Essentiality, while Jaccard measures agreement on Completeness and Determinateness, respectively. Stage 2 transformation accuracy is applicable only to LogicScore.}
  \label{tab:human_logicscore_llm_judge}
\end{table}

\begin{figure*}[t]
    \centering
    \includegraphics[width=0.8\linewidth]{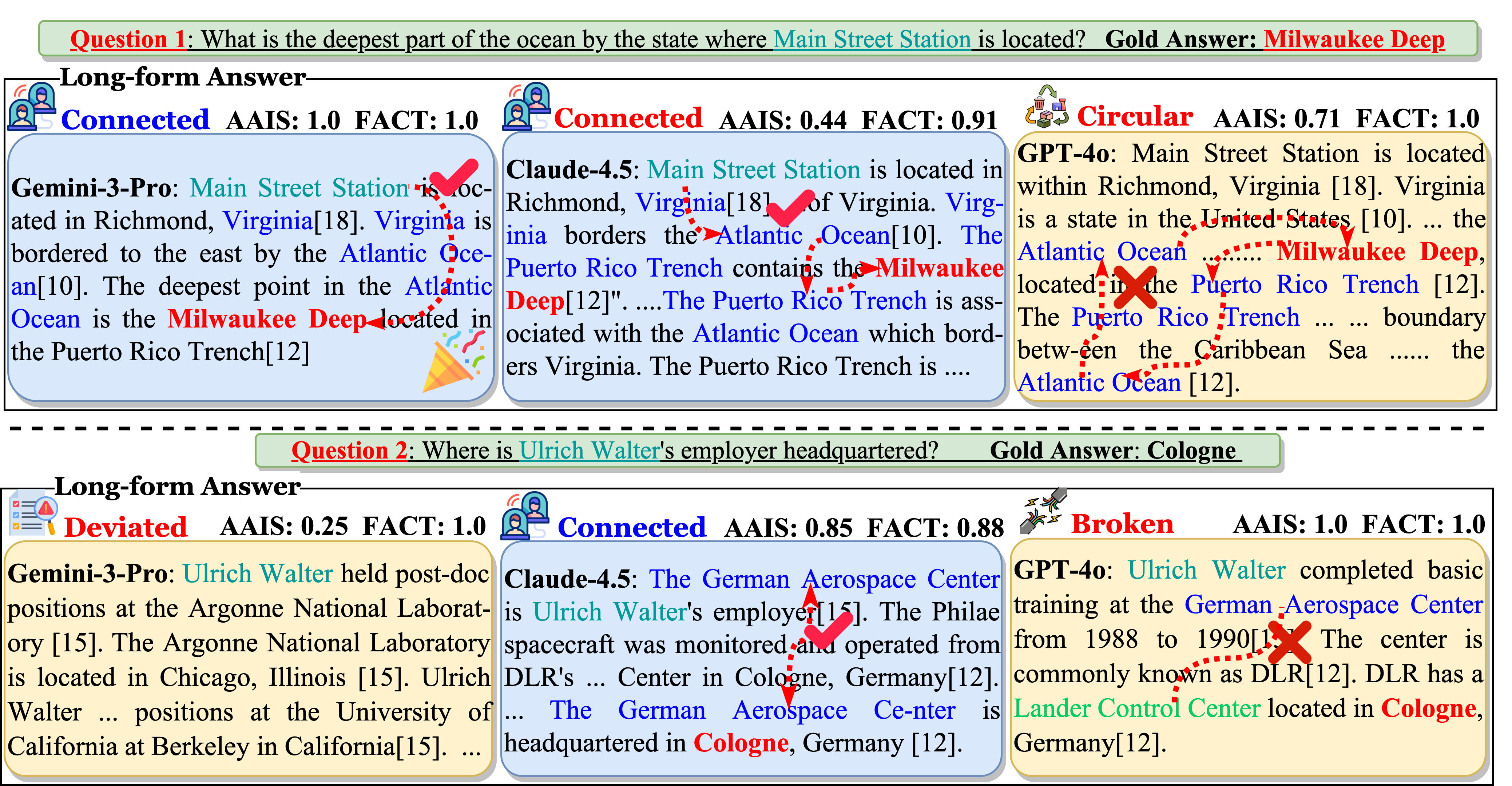}
    \caption{We observe three logic error types when prompting LLMs to generate attributed long-form answers: \textcolor{red}{Circular} denotes self-referential reasoning; \textcolor{red}{Deviated} represents a fundamental divergence in the reasoning trajectory; \textcolor{red}{Broken} signifies a logical discontinuity in the deductive chain. The \includegraphics[height=0.8em]{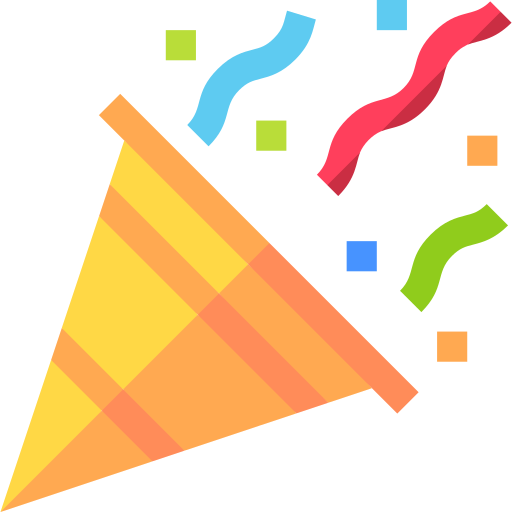} symbolizes the ideal reasoning paradigm.}
    \label{fig:case}
\end{figure*}

\label{differe_llm}
\begin{table}[t] 
    \centering
    \small
    \setlength{\tabcolsep}{0.15cm}{
    \begin{tabular}{lcccccc}
    \toprule
    \multirow{2}{*}{Models} & \multicolumn{3}{c}{GPT-5-mini} & \multicolumn{3}{c}{GPT-5-nano} \\ 
    \cmidrule(lr){2-4}  \cmidrule(lr){5-7} 
     & Esse. & Comp. & Dete. & Esse. & Comp. & Dete.\\ 
    \midrule
    GPT-4o & 79.91 & 82.75 & 91.24 & 74.08 & 78.14 & 93.22\\
    Gemini-3-pro & 70.47 & 75.10 & 92.35 & 74.14 & 70.27 & 94.15\\
    Claude-4.5 & 79.63 & 82.11 & 93.44 & 75.02 & 79.22 & 94.35\\ 
    \bottomrule
    \end{tabular}%
    }
    \caption{Cohen's Kappa scores across backbone LLMs. Ranges 0.61–0.80 and $>$ 0.80 indicate substantial and perfect agreement \citep{mchugh2012interrater}. We replace the original LLM used in \textsc{LogicScore} with GPT-5-mini and GPT-5-nano.}
    \label{tab:differe_llm}
\end{table}

\subsection{Analysis of Partial Reasoning Utility}

To provide a finer-grained interpretation of responses for which no complete question-to-answer path is recovered, we analyze Partial Reasoning Utility. Assessing partial reasoning is challenging because a proposition may provide relevant evidence or an intermediate link without completing the answer path. Given the question, gold answer, and extracted propositions, the judge identifies those making relevant and non-redundant progress toward the answer. For each response, Partial Reasoning Utility is the proportion of such propositions among all extracted propositions.

\begin{table}[h]
    \centering
    \small
    \setlength{\tabcolsep}{8pt}{
    \begin{tabular}{lccc}
      \toprule
      Dataset & Claude-4.5 & Gemini-3-Pro & GPT-5.1 \\
      \midrule
      MuSiQue  & 49.78 (510) & 76.20 (567) & 50.81 (562) \\
      HotpotQA & 63.47 (370) & 80.38 (481) & 68.78 (363) \\
      \bottomrule
    \end{tabular}
    }
    \caption{Mean instance-level Partial Reasoning Utility (\%) assessed by GPT-5-mini. Numbers in parentheses denote the number of evaluated responses.}
    \label{tab:partial-reasoning-utility}
\end{table}

As shown in Table \ref{tab:partial-reasoning-utility}, Gemini-3-Pro achieves the highest Partial Reasoning Utility at 76.20\%, compared with 49.78\% for Claude-4.5 and 50.81\% for GPT-5.1. These results reveal different degrees of useful intermediate progress across the evaluated models when a complete reasoning path is not recovered.

\section{Case Study}

Figure \ref{fig:case} presents illustrative examples of recovered path structures, together with factual grounding scores from AutoAIS \citep{zhang2025longcite} and FACTSCORE \citep{min2023factscore}. We report AutoAIS F1 (AAIS) computed from its precision and recall.

\textbf{Path Failures.} The first case illustrates a \textcolor{red}{\textbf{Circular}} pattern: the recovered chain returns to \texttt{Milwaukee Deep} rather than connecting the question to its answer. For the second question, Gemini-3-Pro shows a \textcolor{red}{\textbf{Deviated}} path containing steps that do not connect to the required multi-document relation, whereas GPT-4o exhibits a \textcolor{red}{\textbf{Broken}} path with an unresolved link between the German Aerospace Center and the Lander Control Center. These examples illustrate distinct ways in which an entity-linked path may fail to connect a question to its answer.

\textbf{Completeness and Essentiality.} Claude-4.5 recovers a complete path for Question 1 but includes additional propositions outside the recovered path. Its Essentiality is therefore lower despite successful path recovery. Similarly, Question 2 obtains an Essentiality of 33.3\%. The difference between the structural diagnostics and factual scores, for example, FACT 91\% versus AAIS 44\% in Question 1, illustrates that factual grounding and recovered reasoning structure capture complementary properties of long-form answers.

\textbf{Ideal Paradigm.} As shown in Figure \ref{fig:case} (\includegraphics[height=1em]{confetti.png}), Gemini-3-Pro exemplifies the ideal reasoning paradigm on Question 1. It exhibits the optimal deductive path: \texttt{Milwaukee Deep} $\to$ \texttt{Virginia} $\to$ \texttt{Atlantic Ocean} $\to$ \texttt{Main Street Station}. By generating the minimum necessary reasoning steps alongside precise attribution formatting, the model achieves superior factual accuracy without redundancy.

\section{Conclusion and Limitations}
In this paper, we introduce \textsc{LogicScore}, a unified evaluation framework that shifts the paradigm from local factual verification to global reasoning scrutiny. Our extensive analysis of over 20 LLMs reveals that high factual attribution often masks systemic reasoning failures, a phenomenon termed \textit{factual myopia}. 
By formalizing long-form answers into Horn Rules, we establish that resolving this \textit{myopia} requires models to prioritize both deductive integrity (\textit{Completeness}) and expression efficiency (\textit{Essentiality}). 
Ultimately, we identify the optimization of \textit{logical density} as a critical frontier for next-generation reasoning models. 

Our framework relies on a Horn-inspired, entity-linked representation, which may not capture the full spectrum of open-domain and non-linear reasoning structures. To bridge this gap, future research could extend the framework with richer and more complex task-specific representations while preserving its focus on reasoning coherence alongside factual grounding.

\bibliography{aaai2027}


\newpage

\clearpage

\appendix

\end{document}